\newcommand{\CLASSINPUTtoptextmargin}{23mm}
\newcommand{\CLASSINPUTbottomtextmargin}{19mm}
\newcommand{\CLASSINPUTinnersidemargin}{19mm}
\newcommand{\CLASSINPUToutersidemargin}{19mm}
\documentclass[letterpaper, 10 pt]{ieee_templates/IEEEtran}
\usepackage{times}
\usepackage[T1]{fontenc}
\usepackage[english]{babel}
\usepackage{amsmath,amsfonts,amssymb,amsthm}
\usepackage{wrapfig}
\usepackage[linesnumbered,ruled,vlined]{algorithm2e}

\usepackage{graphicx}
\usepackage{float}
\usepackage{microtype}
\usepackage{color}
\usepackage[numbers]{natbib}

\usepackage{multicol}
\usepackage[bookmarks=true]{hyperref}
\usepackage[caption=false]{subfig}
\usepackage[percent]{overpic}
\usepackage{tikzscale}
\usepackage{pgfplots}
\pgfplotsset{compat=1.16}
\usepackage{tikz}
\usepackage{mathtools}
\usepackage{wrapfig}
\usepackage{bbold}

\renewcommand{\abovecaptionskip}{0pt}
\newcommand{\CVaR}{\mathrm{CVaR}}
\newcommand{\VaR}{\mathrm{VaR}}

\usepackage{etoolbox}
\makeatletter
\patchcmd{\@makecaption}
  {\scshape}
  {}
  {}
  {}
\makeatother

\usepackage{setspace}

\IEEEoverridecommandlockouts 

\begin{document}
\title{\vspace{2mm}Learning Risk-aware Costmaps for Traversability in Challenging Environments
}

\author{David D. Fan$^{1,2}$, Sharmita Dey$^{1,3}$, Ali-akbar Agha-mohammadi$^{2}$,
and Evangelos A. Theodorou$^{1}$
\thanks{This paper was recommended for publication by Editor 
Jens Kober upon evaluation of the Associate Editor and Reviewers' comments.}
\thanks{$^{1}$David D. Fan and Evangelos A. Theodorou are with the Institute for Robotics and Intelligent Machines, Georgia Institute of Technology, Atlanta, GA, USA {\tt\footnotesize contact@daviddfan.com}}%
\thanks{$^{2}$David D. Fan, Sharmita Dey and Ali-akbar Agha-mohammadi are with NASA Jet Propulsion Laboratory, California Institute of Technology, Pasadena, CA, USA}%
\thanks{$^{3}$Sharmita Dey is with the University of Goettingen, Germany}%
}


\markboth{IEEE International Conference on Robotics and Automation, 2022}
{Fan \MakeLowercase{\textit{et al.}}: Learning Risk-aware Costmaps for Traversability in Challenging Environments} 
\maketitle


\begin{abstract}
One of the main challenges in autonomous robotic exploration and navigation in unknown and unstructured environments is determining where the robot can or cannot safely move.  A significant source of difficulty in this determination arises from stochasticity and uncertainty, coming from localization error, sensor sparsity and noise, difficult-to-model robot-ground interactions, and disturbances to the motion of the vehicle.  Classical approaches to this problem rely on geometric analysis of the surrounding terrain, which can be prone to modeling errors and can be computationally expensive.  Moreover, modeling the distribution of uncertain traversability costs is a difficult task, compounded by the various error sources mentioned above.  In this work, we take a principled learning approach to this problem.  We introduce a neural network architecture for robustly learning the distribution of traversability costs.  Because we are motivated by preserving the life of the robot, we tackle this learning problem from the perspective of learning tail-risks, i.e. the conditional value-at-risk (CVaR).  We show that this approach reliably learns the expected tail risk given a desired probability risk threshold between 0 and 1, producing a traversability costmap which is more robust to outliers, more accurately captures tail risks, and is more computationally efficient, when compared against baselines.  We validate our method on data collected by a legged robot navigating challenging, unstructured environments including an abandoned subway, limestone caves, and lava tube caves.
\end{abstract}

\IEEEpeerreviewmaketitle


\section{Introduction}
\begin{figure}
  \centering
  \includegraphics[width=\linewidth]{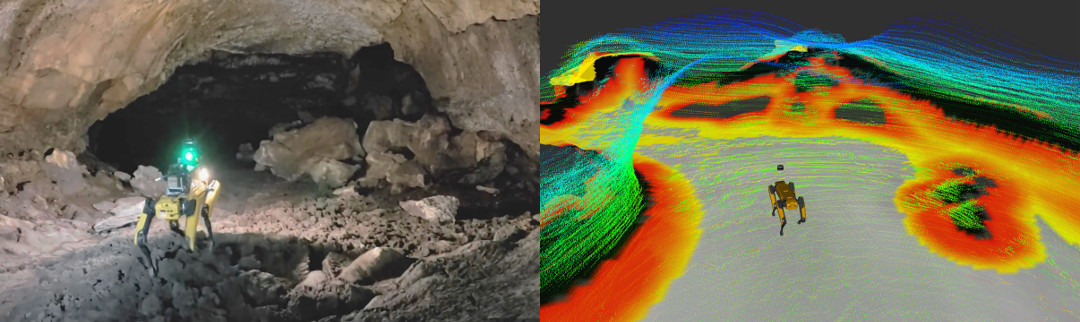}
  \caption{Left:  Spot autonomously exploring Valentine Cave, Lava Beds National Monument, CA, USA.  Right:  LiDAR pointcloud and computed costmap in the same environment.  In this work, we aim to infer a CVaR costmap from the LiDAR pointcloud.}
  \label{fig:splash}
\end{figure}

\IEEEPARstart{U}{ncertainty} is ever-present in robotic sensing and navigation.  Localization error can severely degrade the quality of a robot's environment map, leading to a robot over-estimating or under-estimating the safety of traversing some particular terrain \cite{thakker2021autonomous}.  Sensor noise, sparsity, and occlusion can similarly degrade localization and mapping performance, especially in perceptually degraded environments such as in dark, dusty, or featureless environments \cite{hero2019isrr}.  Moreover, the environment itself is a constant source of mobility uncertainty, as ground-vehicle interactions are notoriously difficult to model and even more difficult to accurately compute in real time - especially in unstructured environments such as those filled with slopes, rough terrain, low traction, rubble, narrow passages, and the like (Figure \ref{fig:splash}) \cite{kalita2018path}.  In each case, modeling these uncertainties often relies on computationally tractable but distributionally restrictive assumptions; for example, assuming Gaussian distributions for localization error, or worst-case bounds on a settled pose \cite{otsu2020fast}.  

We wish to more accurately quantify the uncertainty of computing traversability costs.  This allows the robot to have some measure of control over the level of risk it is willing to take.  This control can be used by task or mission-level planners which weigh a variety of factors to decide the best course of action \cite{kim2021plgrim}.

Risk measures, often used in finance and operations research, provide a mapping from a random variable (usually the cost) to a real number.  These risk metrics should satisfy certain axioms in order to be well-defined as well as to enable practical use in robotic applications \cite{majumdar2020should}.  Conditional value-at-risk (CVaR) is one such risk measure that has this desirable set of properties, and is a part of a class of risk metrics known as \textit{coherent risk measures} \cite{artzner1999coherent}.

Accurately constructing these risk measures poses a challenge, often relying on distributional assumptions such as Gaussianity \cite{otsu2020fast}.  In our previous work \cite{fan2021step}, we model traversability risks as Gaussian random variables and use them to efficiently construct risk measures which are useful for kinodynamic planning.  While these approaches have the advantage of allowing for efficient, kinodynamic, risk-aware planning, they may suffer from relying too heavily on these modeling assumptions.  Such assumptions include:  1) Localization error is accurately known (and is Gaussian), 2) various component risks can be accurately determined from a map (such as slope, step size, roughness, collision), and 3) these risks are normally distributed and both mean and variances can be accurately modeled.

Learning CVaR from data is an attractive approach to bypassing these complex issues.  Since \citet{koenker2001quantile} showed that minimizing a \textit{quantile} loss function results in unbiased prediction of the VaR, or quantile function, many deep learning applications of this idea have been examined \cite{kivaranovic2020adaptive, dabney2018distributional}.  However, there is little work on predicting CVaR directly, instead it is common to make assumptions of the underlying distribution.  More recently, \citet{Peng2021} propose a method for predicting CVaR with a deep neural network under assumptions of i.i.d. samples, and prove convergence of the scheme.  In this work, we propose an approach for distribution-free CVaR learning which scales to images and large datasets.  We construct loss and evaluation functions for both VaR and CVaR which enforce monotonicity with respect to the risk probability.

Learning-based costmaps have seen more recent development for robotics and autonomous driving.  The concept is attractive, possibly because it bypasses traditional modeling approaches, reduces computation, and improves performance from data.  Recent work on CVaR based costmaps rely on assumptions of distribution (Gaussian) or sampling to estimate CVaR \cite{choudhry2021cvar}.  \citet{hakobyan2021distributionally} learns an approximating CVaR costmap from precomputed CVaR values on a collision-avoidance task.  Our contribution to this space is to propose learning CVaR directly in a distribution-free manner, with a novel application to navigation in challenging terrain.

We propose an architecture in which raw or minimally processed pointcloud data is transformed and fed into a convolutional neural network (CNN).  In contrast to other approaches, the network directly produces a CVaR costmap which encodes the traversability risk, given a desired probability of confidence.  Note that in this work we treat the risks in each cell in the costmap as \textit{static} and independent of the others, that is, we do not consider the risk of traversing multiple cells sequentially.  (See recent work \cite{fan2021step, majumdar2020should} which addresses the notion of dynamic risk metrics in the context of planning and assessing risk over a path).  We restrict our approach to assessing point-wise risks which can then be used to commit the robot to finding a path which uses these risks as a constraint, avoiding the riskiest regions and thereby remaining safe.  (In contrast, in \cite{fan2021step} we minimize the CVaR of the risk along an entire path.)  
 
Our contributions are summarized as follows:
\begin{itemize}
    \item A novel neural network architecture for transforming pointcloud data into risk-aware costmaps.
    \item A loss function which trains this network to produce quantile and CVaR values without distribution assumptions.
    \item A solution to a challenging traversability learning task in unknown environments, validated on a wide range of unstructured and difficult terrain.
    \item Evaluation metrics which assess the goodness-of-fit of the network and comparison between different baseline approaches.
\end{itemize}

\section{Method}
In this section we establish definitions and describe our method for learning CVaR costmaps.  We first formally define our notion of "traversability".  Then we discuss risk metrics, value-at-risk (VaR) and conditional value-at-risk (CVaR).  We discuss the losses used to train a network to produce these values, and describe how we obtain training labels.
\subsection{Traversability as a random variable}
We define \textit{traversability} as the capability for a ground vehicle to reside over a terrain region under an admissible state \cite{papdakis2013survey}.  We represent traversability as scalar value $r$ which indicates the magnitude of damage (or cost of repair) the robot will experience if placed in that state.  We then wish to construct a model of this traversability given the robot's state and knowledge:
\begin{align}
    R = \mathcal{R}(m, x)
\end{align}
where $x\in\mathcal{X}$ is the robot position in 2-D coordinates, $m\in\mathcal{M}$ represents the current belief of the local environment, and $\mathcal{R}(\cdot)$ is a traversability assessment model.  This model captures various unfavorable events such as collision, getting stuck, tipping over, high slippage, to name a few. A mobility platform has a unique assessment model which reflects its mobility capability.  In most real-world applications where perception capabilities are limited, the true value can be highly uncertain.  To handle this uncertainty, consider a map belief, i.e., a probability distribution $p(m | x_{0:k}, z_{0:k})$ over a possible set $\mathcal{M}$, where $z_{0:k}$ are sensor observations.  Then, the traversability estimate is a random variable $R: (\mathcal{M} \times \mathcal{X}) \longrightarrow \mathbb{R}$.  We call this probabilistic mapping from map belief and state to possible traversability cost values a \textit{risk assessment model}.

\subsection{Risk Metrics, VaR and CVaR}
A risk metric $\rho(R):R\rightarrow\mathbb{R}$ is a mapping from a random variable to a real number which quantifies some notion of risk.  Local environment information is encoded within $m$.  This can be raw sensor data, e.g. observations, or processed map data (e.g. represented as $m = (m^{(1)}, m^{(2)}, \cdots)$ where $m^{i}$ is the $i$-th element of the map).  In the case of a 2-D or 2.5-D representation, $i$ represents the cell index.  Risk metrics can be coherent, which mean they satisfy six properties with respect to the random variables they act upon, namely: normalized, monotonic, sub-additive, positive homogeneity, and translation invariant.  \citet{majumdar2020should} argues that the consideration of risk in robotics should utilize coherent risk metrics for their intuitive and well-formulated properties.  
In this work we are concerned with right-tail risk only, since the random variable $R$ is a positive "cost" for which we seek to avoid high values.

Let $\alpha \in [0, 1]$ denote the \textit{risk probability level}.  High values of $\alpha$ imply more risk. The value-at-risk of the random variable $R(m,x)$ can be defined by (we write $R$ for brevity):
\begin{align}
    \VaR_{\alpha}(R)\coloneqq
    & \mathrm{inf}\{z \in \mathbb{R} | P(R < z) > \alpha\} \label{eq:var}
\end{align}
This is also known as the $\alpha$-quantile.  While there are multiple ways to define the conditional value-at-risk, one common definition is as follows:
\begin{equation}
    \CVaR_{\alpha}(R)\coloneqq\mathbb{E}[R|R \geq \VaR_{\alpha}(R)].
\end{equation}


\begin{figure}
    \centering
        \resizebox{0.4\textwidth}{!}{%
    \begin{tikzpicture}
    \tikzstyle{every node}=[font=\large]
    \pgfmathdeclarefunction{gauss}{2}{%
      \pgfmathparse{1000/(#2*sqrt(2*pi))*((x-.5-8)^2+.5)*exp(-((x-#1-6)^2)/(2*#2^2))}%
    }
    
    \pgfmathdeclarefunction{gauss2}{3}{%
    \pgfmathparse{1000/(#2*sqrt(2*pi))*((#1-.5-8)^2+.5)*exp(-((#1-#1-6)^2)/(2*#2^2))}
    }
    
    \begin{axis}[
      no markers, domain=0:16, range=-2:8, samples=200,
      axis lines*=center, xlabel=$R$, ylabel=$p(R)$,
      every axis y label/.style={at=(current axis.above origin),anchor=south},
      every axis x label/.style={at=(current axis.right of origin),anchor=west},
      height=5cm, width=17cm,
      xtick={0,5.5,7,10}, ytick=\empty,
      xticklabels={$0$, , , ,},
      enlargelimits=true, clip=false, axis on top,
      grid = major
      ]
      \addplot [fill=cyan!20, draw=none, domain=7:15] {gauss(1.5,2)} \closedcycle;
      \addplot [very thick,cyan!50!black] {gauss(1.5,2)};
     
     \pgfmathsetmacro\valueA{gauss2(5.5,1.5,2)}
     \draw [gray] (axis cs:5.5,0) -- (axis cs:5.5,\valueA);
      \pgfmathsetmacro\valueB{gauss2(10,1.5,2)}
      \draw [gray] (axis cs:4.5,0) -- (axis cs:4.5,\valueB);
        \draw [gray] (axis cs:10,0) -- (axis cs:10,\valueB);
     
     
     \draw [gray] (axis cs:1,0)--(axis cs:5.5,0);
     \node[below] at (axis cs:7.5, -0.1)  {$\mathrm{VaR}_{\alpha}(R)$}; 
    \node[below] at (axis cs:5.5, -0.1)  {$\mathbb{E}(R)$}; 
    \node[below] at (axis cs:10.5, -0.1)  {$\mathrm{CVaR}_{\alpha}(R)$};
    \draw [yshift=2cm, latex-latex](axis cs:7,0) -- node [fill=white] {Probability~$1-\alpha$} (axis cs:16,0);
    \end{axis}
    \end{tikzpicture}
    }
    \caption{Comparison of the mean, VaR, and CVaR for a given risk level $\alpha \in (0,1]$. The axes denote the values of the stochastic variable $R$, which in our work represents traversability cost. The shaded area denotes the $(1-\alpha)\%$ of the area under $p(R)$.  $\CVaR_{\alpha}(R)$ is the expected value of $R$ under the shaded area. (From \cite{fan2021step}).}
    \label{fig:cvar}
\end{figure}
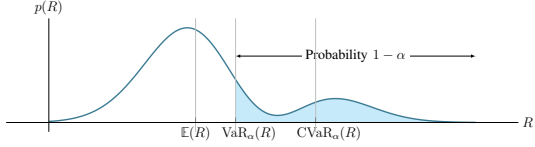

\subsection{Learning VaR and CVaR}
We wish to construct a model with inputs $m$, $x$, and $\alpha$, and outputs $\CVaR_\alpha(R(m,x))$, parameterized by network weights $\theta$.  We denote this model as $C_\theta(m,x,\alpha)$.  To learn CVaR we take a joint approach where we learn both VaR and CVaR together.  We construct a similar VaR model, also as a function of $m$ and $x$ as $\VaR_\alpha(R(m,x))$, which we denote $V_\theta(m,x,\alpha)$.  Learning VaR can be accomplished by minimizing the Koenker-Bassett error with respect to $\theta$ \cite{koenker2001quantile}:
\begin{multline}
    l^V_{\alpha}(\theta) = \alpha(R(m,x)-V_{\theta}(m,x,\alpha))_{+} \\
    + (1-\alpha)(R(m,x)-V_{\theta}(m,x,\alpha))_{-}
\end{multline}
where $(\cdot)_{+}=\max(\cdot, 0)$ and $(\cdot)_{-}=\min(\cdot, 0)$.  From the estimated VaR, we can compute the expected CVaR and construct an L1-loss for $C_\theta$:
\begin{align}
    l_\alpha^C(\theta) &= |C_\theta(m,x,\alpha) - R(m,x)|\mathbb{1}_{R(m,x) \geq V_\theta(m,x,\alpha)}
\end{align}
With the assumption of i.i.d. sampled data, when $l^V_{\alpha}(\theta)$ and $l^C_{\alpha}(\theta)$ are minimized, $V_\theta$ will approximate $\VaR$ and $C_\theta$ will approximate $\CVaR$.  If during training we uniformly randomly sample values of $\alpha\in[0,1]$, then the input $\alpha$ to these models should be meaningful as well.  

In practice, a few modifications to these losses are needed to improve numerical stability.  First, similar to \cite{dabney2018distributional}, we smooth the quantile loss function $l^V_\alpha$ near the inflection point $V_\theta=R$ by using a modified Huber loss \cite{huber1992robust}:
\begin{equation}
    l_h(e, \alpha) = 
    \begin{cases}
        (1 - \alpha) |e| & \mathrm{if} ~ e \leq \frac{-h}{1-\alpha}\\
        \frac{1}{2h}((1 - \alpha) |e|)^2+\frac{h}{2} & \mathrm{if}~ \frac{-h}{1-\alpha} < e \leq 0 \\
        \frac{1}{2h}(\alpha|e|)^2+\frac{h}{2} & \mathrm{if} ~0 < e \leq \frac{h}{\alpha} \\
        \alpha |e| & \mathrm{if} ~  \frac{h}{\alpha} < e
    \end{cases}
\end{equation}
where $h\in \mathbb{R}^{+}$ controls how much smoothing is added.  The new VaR loss is then:
\begin{equation}
    \hat{l}_\alpha^V(\theta) = l_h(R(m,x)-V_\theta(m,x,\alpha), \alpha).\label{eq:var_huber}
\end{equation}


Second, instead of learning $C_{\theta}$ directly, we can learn the residual between $C_{\theta}$ and $V_{\theta}$.  Suppose we construct a model $\hat{C}_\theta$ instead of $C_\theta$, and let $C_\theta = V_\theta + \hat{C}_\theta$.  Then the loss $l_\alpha^C(\theta)$ can be written as:
\begin{align}
    l_\alpha^C(\theta) &= |\hat{C}_\theta - (R(m,x) - V_\theta(m,x,\alpha))|\mathbb{1}_{R \geq V_\theta}\label{eq:cvar_modified}
\end{align}
This serves to separate the error signals between the two models.

Third, we introduce a monotonic loss to enforce that increasing values of $\alpha$ result in increasing values of $V_\theta(m,x,\alpha)$ and $C_\theta(m,x,\alpha)$.  This can be done by penalizing negative divergence of the output with respect to $\alpha$ \cite{fan2020deep}.  We also introduce a smoothing function to prevent instability near the inflection point when the divergence equals 0.  The total monotonic loss is:
\begin{align}
    d_V &= (\nabla_\alpha V_\theta(m,x,\alpha))_{-}\\
    d_C &= (\nabla_\alpha C_\theta(m,x,\alpha))_{-}\\
    s(d) &= \exp(d) - d - 1.0\\
    l^m(\theta) &= s(d_V) + s(d_C)\label{eq:monotonic}
\end{align}
In practice we find that under gradient-based optimization, this loss decreases to near 0 in the first epoch and does not noticeably affect the minimization of the quantile and CVaR losses.

To summarize, the total loss function is the sum of the modified Huber quantile loss (\ref{eq:var_huber}), the L1 residual CVaR loss (\ref{eq:cvar_modified}), and the monotonic loss (\ref{eq:monotonic}):
\begin{equation}
L(\theta; R,m,x,\alpha) = \lambda_V \hat{l}_\alpha^V(\theta) + \lambda_C l_\alpha^C(\theta) + \lambda_m l^m(\theta) \label{eq:loss}
\end{equation}

\begin{figure}
  \centering
  \includegraphics[width=0.4\textwidth]{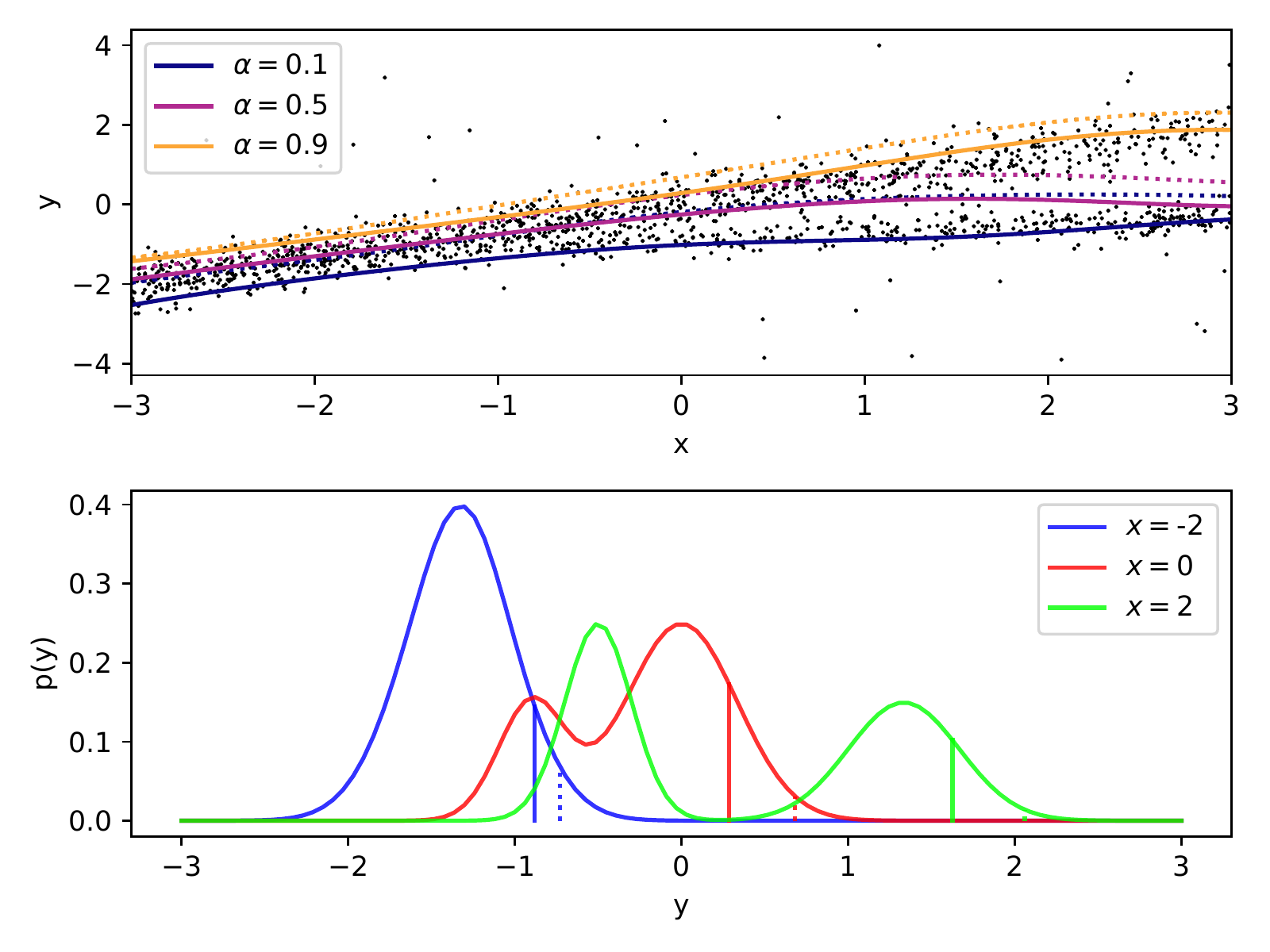}
  \caption{Learned VaR and CVaR on a toy 1-D problem.  Top:  Samples drawn from a distribution $y\sim f(x)$ which is multimodal and heteroskedastic.  Solid and dotted lines show learned VaR and CVaR levels, respectively, for different values of $\alpha$.  Bottom:  PDF of the true distribution for varying values of $x$.  Also marked are the learned VaR (solid vertical line) and CVaR (dotted vertical line) values, when $\alpha=0.9$. }
  \label{fig:toy_cvar}
\end{figure}

We minimize this loss over a dataset, sampled i.i.d from the distribution of $R$, which is a function of $m$ and $x$.  We also sample from a uniform distribution for $\alpha$.  In other words, we seek to minimize the expected loss with respect to $\theta$:
\begin{align}
    \mathbb{E}_{R,\alpha}&[L(\theta;R,m,x,\alpha)] \\
    =&\int L(\theta|R,m,x,\alpha)p(R,m,x)p(\alpha)dRd\alpha \nonumber\\
    =&\int L(\theta|R,m,x,\alpha)p(R|m,x)p(m,x)p(\alpha)dRdmdxd\alpha \nonumber
\end{align}
This expectation is approximated using stochastic gradient descent \cite{shamir2013stochastic}.  Given the dataset $\mathcal{D}=\{m_n,x_n\}_n$, the  distribution $p(m,x)$ is sampled from the distribution of data collected by the robot as it moves in the environment, observing samples of its position $x$ and its environment data $m$.  The distribution $p(R|m,x)$ is sampled by determining the traversability cost given a sampled $(m,x)$.  This determination may be stochastic or imprecise, and may possibly come from any traversability assessment method.  Happily, our learning approach should capture this stochasticity.  In Figure \ref{fig:toy_cvar}, we demonstrate this learning approach on a toy 1-D problem.  We use a 3 layer feed-forward MLP neural network to learn the state-varying VaR and CVaR values of a random variable. We notice that our learning approach captures the risk of the known distributions accurately.  Next, we describe the application of this loss function to learning unknown distributions of 2-D traversability costs from 3D pointclouds.

\subsection{Obtaining Ground Truth Labels}
As mentioned, computing traversability costs from sensor data is a rich field in itself.  Computing these costs may include geometric analysis \cite{gestalt}, semantic understanding and detection \cite{kostavelis2015semantic}, proprioceptive sensing \cite{lew2019contact}, or hand-labeling.  Our approach is extensible to both dense and sparse ground truth labels in the costmap.  In this work we focus on costs arising through geometric analysis.  We leverage prior work (\citet{fan2021step}), which itself follows a tradition of geometric analyses for ground vehicle traversal \cite{otsu2020fast, gestalt}.  These analyses are based on pointcloud data and provide model-based costs which are computed in a dense local region around the robot, but are affected by sensor occlusion, sparsity, noise, and localization error.  These sources of ambiguity and stochasticity are exactly what we aim to capture with our risk-aware model.  However, instead of having to explicitly model these uncertainties, we simple compute a mean value best-guess at the traversability cost, and use learning to aggregate the variation of these costs.

\section{Implementation Details}
In this section, we discuss details of the dataset, data processing, network architecture, and training procedures.

\subsection{Dataset}
\begin{figure*}
  \centering
  \includegraphics[width=1.0\textwidth]{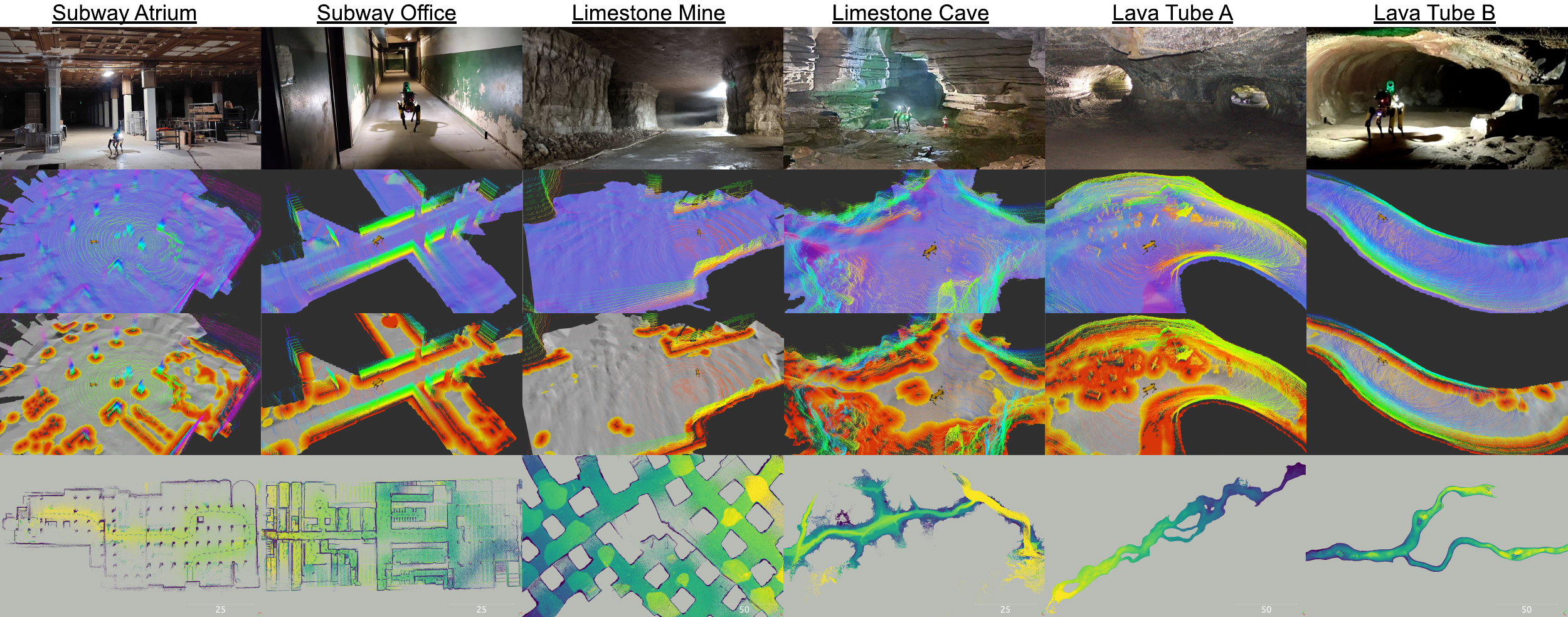}
  \caption{Datasets collected in 6 different environments.  Top row:  Photo of the environment.  Second row:  LiDAR pointcloud and elevation map produced after ground segmentation.  Third row:  Handcrafted risk map with varying risk (white: safe ($r<=0.1$), yellow to red: moderate ($0.1 < r <= 0.9$), black: risky ($r > 0.9$)).  Pointclouds are also shown.  Bottom row:  Map of the entire environment, generated by aggregating LiDAR pointclouds during each data collection run.  Scale (in meters) is shown in the lower right corner.}
  \label{fig:dataset}
\end{figure*}

Our motivation for tackling this problem comes from extensive field testing in various underground caves and decaying urban environments \cite{agha2021nebula}.  We collected data from autonomous exploration runs in six different environments of various types (Figure \ref{fig:dataset}, Table \ref{table:datasets}).  These environments are as follows:  
\begin{itemize}
    \item An abandoned subway station in Los Angeles, CA, the first floor has a large open atrium with pillars (Subway Atrium / SA).
    \item The basement floor of the same subway station, featuring offices and narrow corridors (Subway Office / SO).
    \item Kentucky Underground Storage in Lexington, KY, which consists of very large cavernous grids of tunnels (Limestone Mine / LM)
    \item Wells Cave in Pulaski County, KY, which is a natural limestone cave with very rough floors, narrow passages, low ceilings, and rubble (Limestone Cave / LC)
    \item Valentine Cave at Lava Beds National Park, CA, which is an naturally formed ancient lava tube, with sloping walls and rough, rocky floors (Lava Tube A / TA)
    \item Mammoth Cave at Lava Beds National Park, CA, which is also a lava tube but with sandy, sloping floors and occasional large piles of rubble (Lava Tube B / TB).
\end{itemize}

\begin{table}[h!]
\centering
\tiny
\vspace*{0.1cm}
\begin{tabular}{c|c|c|c|c|c|c} 
  & SA & SO & LM & LC & TA & TB \\
  \hline
\# Samples & 1585 & 2883 & 942 & 331 & 852 & 1148 \\
Duration (min) &	53	&96	&31	&11	&28	&38 \\
Distance (m) 	&1000	&1000&	600	&300	&600	&800\\
Min Width (m) 	&5	&1	&10	&0.5	&1	& 3 \\
Max Width (m) 	&50	&10	&20	&5	&10	&10
\vspace*{0.1cm}
\end{tabular}
\caption{Details of datasets, with number of data samples, duration of the runs, approximate distance traveled, and average width of the passages in the environment.}
\label{table:datasets}
\end{table}

Data was collected using Boston Dynamics's Spot legged robot equipped with JPL's NeBula payload \cite{agha2021nebula}.  The payload includes onboard computing, one VLP-16 Velodyne LiDAR sensor, and a range of other cameras and sensors.  Spot is equipped with 5 Intel RealSense depth cameras which are pointed at the ground. A new data sample was added to the dataset at approximately 0.5Hz, while the average top speed of the robot was set to 1m/s.     

\subsection{Computing Traversability Cost}
\label{sec:handcraft}
Traversability costs are computed online and saved along with dataset.  These costs are generated via a risk-aware traversability pipeline \cite{fan2021step}, which we will briefly summarize here.  First, we aggregate LiDAR pointclouds over time, using odometry.  Odometry is generated using a combination of Spot's on-board VIO and ICP-based pointcloud matching.  Next, we perform ground segmentation to isolate "ground" points, "obstacle" points, and "ceiling" points.  We then perform elevation mapping, which probabilistically builds a 2.5-D elevation map using the "ground" points.  Localization noise and delay are taken into account here, resulting in elevation variance information as well.  We then compute costs as a function of various geometric information arising from the elevation map and the pointcloud, such as slope, step size, roughness, negative obstacles, and collision obstacles.  These costs are then aggregated into a combined mean cost (as well as a cost variance).  The traversability cost is scaled between 0 and 1, with 1 being lethal and 0 being safe.  In \cite{fan2021step}, the mean cost and variance of the cost are then used to compute CVaR, with assumption of a Gaussian distribution.  In this work, however, we only use the mean cost which represents $R$.  We aim to directly predict CVaR from the distribution of costs that arise through this analysis and the environment itself.

\subsection{Transforming Pointclouds to Costmaps}
\begin{figure}
  \centering
  \includegraphics[width=\linewidth]{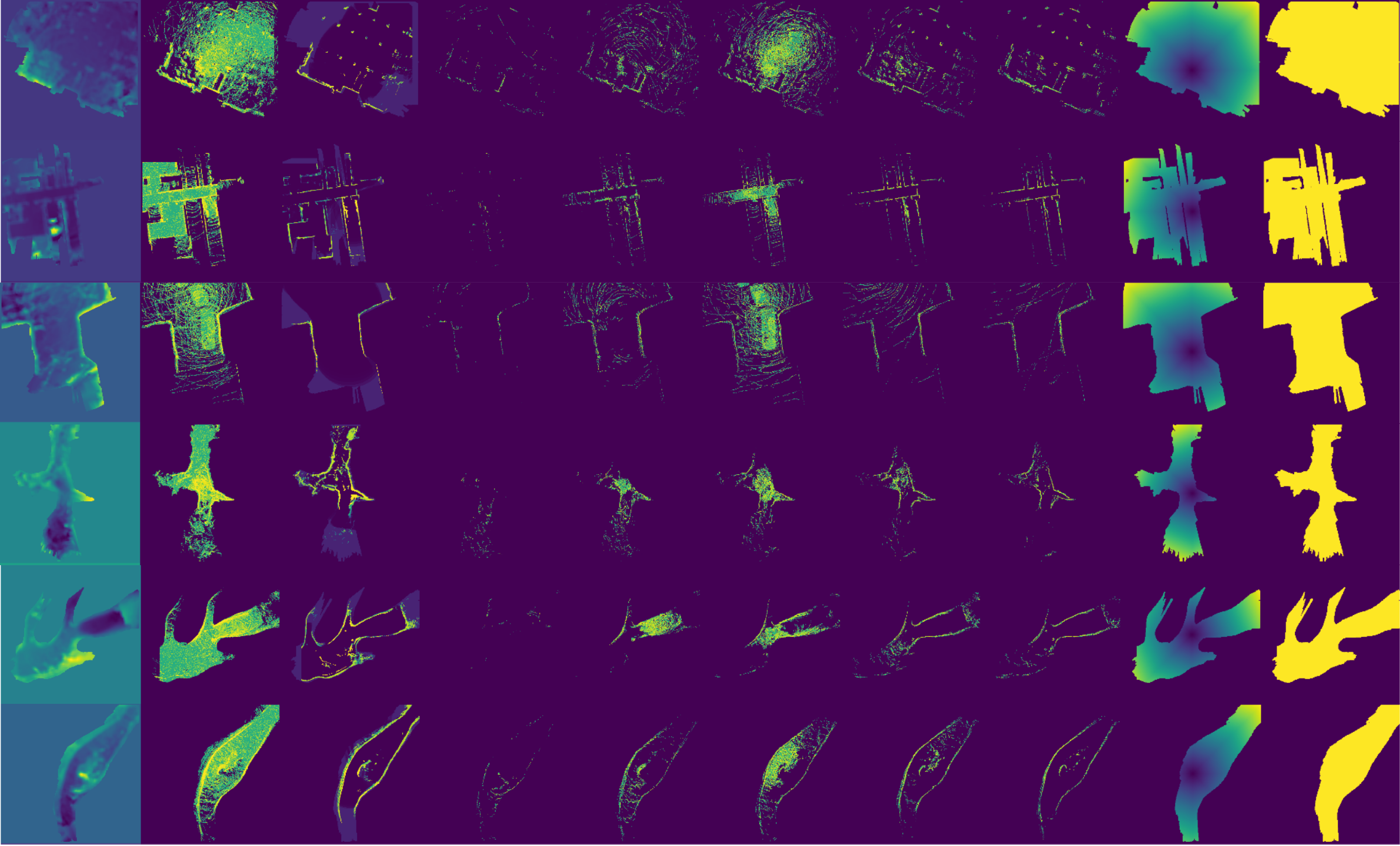}
  \caption{Input features converted from LiDAR pointclouds, showing different features in columns from left to right, while each row corresponds to one sample from each of the 6 datasets.  Features are, from left to right:  1) elevation, 2) number of LiDAR points, 3) obstacle points (older points have a lower intensity), 4-8) number of points in each of 5 z-height bins, relative to the elevation map, with a bin height of 0.1m, 9) distance from the robot location, 10) "known" region mask, which marks regions which have sensor coverage.}
  \label{fig:input_features}
\end{figure}

We aim to construct a network whose inputs are the LiDAR pointcloud, and whose output is a CVaR costmap (2-D).  There are many different approaches for processing LiDAR data using neural networks. We investigated two different architectures.  The first uses an image-to-image autoencoder type of network (namely, U-Net with partial convolutions \cite{liu2018image}), along with a handcrafted histogram binning method for converting LiDAR pointclouds to images.  We bin pointcloud data into 5 height bins relative to the computed elevation map.  We count the number of points which fall into these bins and generate an image from these counts.  We end up with input features where each feature channel is a 2-D image (see Figure \ref{fig:input_features}).  Image sizes are 400x400 pixels, with each pixel representing 0.1m, for a total area of 40m x 40m covered.
We also investigated an approach which uses channelized convolution operations within the network. In this second approach, we convert the unstructured pointclouds into structured voxel grids, with a voxel size of $0.5 m$.  Each voxel contains the averaged intensity values of the points falling into it.  Then, we use a Linknet \cite{chaurasia2017linknet} encoder-decoder architecture with a VGG19 encoder backbone to convert the voxelized pointclouds directly to costmaps. The network was pretrained with Imagenet weights.  While we believe this latter architecture has more expressive power than the first with more weights and layers, we more fully evaluated the first approach, which we felt was more compatible to the size of the datasets which we used for evaluation.

The presence of gaps and occlusions in LiDAR data lead to known and unknown regions in the image-translated input data.  These unknown regions need to be handled by the network.  A naive approach would be to use zero-padding, i.e. replacing unknown pixels with 0.  However, this can lead to blurring and other spurious artifacts at the edges.   For this reason, the use of partial convolutional neural networks lends itself particularly well to this image-to-image learning task with unstructured, unknown masks \cite{liu2018image}.

Additionally, we provide the desired $\alpha$ level as an image to the input of the network.  Since $\alpha$ is an image it can vary spatially.  The loss function takes this desired $\alpha$ into account per-pixel, and trains the network to produce the correct risk level according to the desired input pattern.  During training, we use a uniformly distributed smoothed random pattern, whose distribution spans $\alpha\in[0,1]$.

\begin{figure*}[ht]
  \centering
  \includegraphics[width=\linewidth]{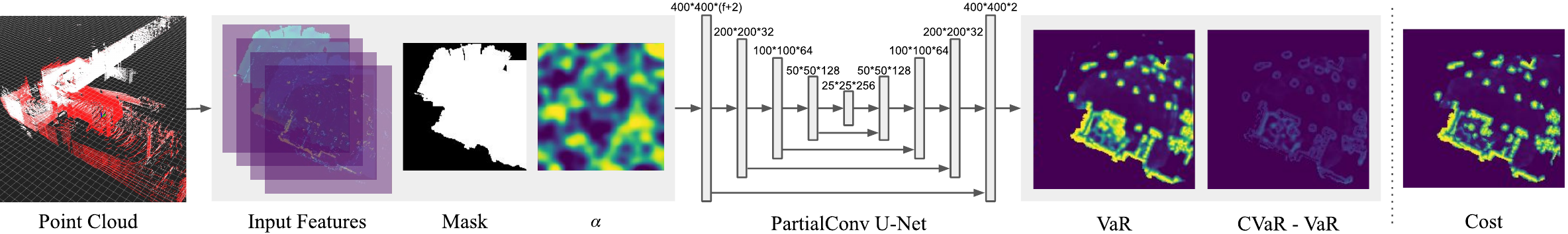}
  \caption{Our pointcloud-to-costmap pipeline.  From left to right:  Raw pointclouds are aggregated and used to create 2-D image-like input features and the mask. A 2-D $\alpha$ channel also provides input to the network.  The PartialConv U-Net architecture maps these input features to 2 output channels, namely $\VaR$ and $\CVaR - \VaR$.  These two outputs are combined with the handcrafted cost labels to compute the loss.}
  \label{fig:arch}
\end{figure*}

Figure \ref{fig:arch} outlines the costmap network architecture used in our approach.  The pipeline consists of three components:  1) Pointcloud-to-image features transform, 2) an image-to-costmap translation network using a PartialConv U-Net, and 3) VaR ($V_\theta(m,x,\alpha)$) and CVaR ($\hat{C}_\theta(m,x,\alpha)$) outputs, which are fed into the loss function (Equation \ref{eq:loss}).  The loss is averaged pixel-wise across the known mask regions only.  This implies that each $\alpha$ value is independent from its neighbors.  



\begin{figure}
  \centering
  \includegraphics[width=\linewidth]{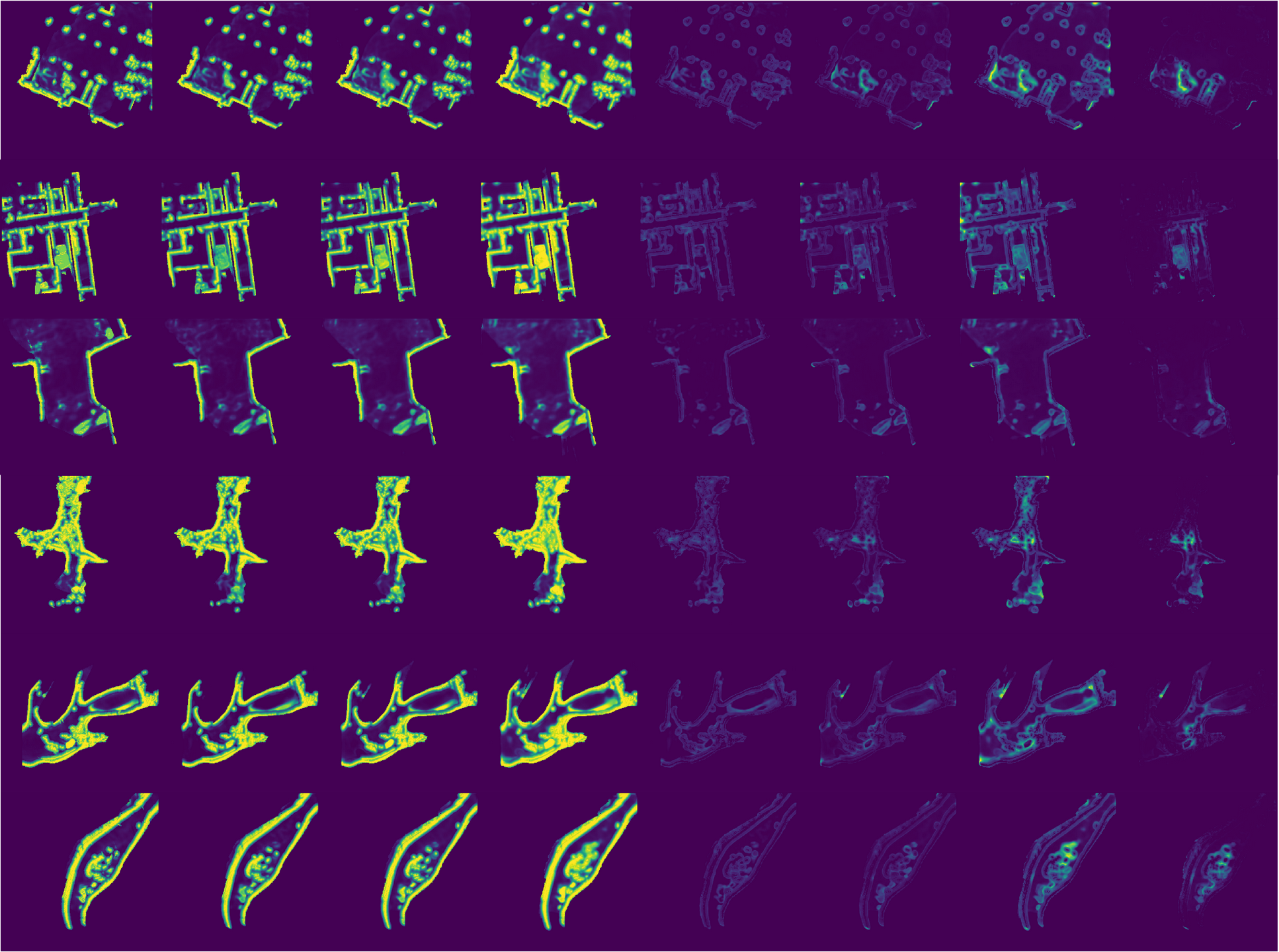}
  \caption{Network outputs on the same input examples shown in Figure \ref{fig:input_features}.  Columns from left to right:  1)  Handcrafted cost label, 2-4) CVaR values with $\alpha=0.1, 0.5, 0.9$ respectively, 5-7) $\CVaR_\alpha - \VaR_{0.1}$ for varying $\alpha=0.1,0.5,0.9$ respectively.  This enables us to more clearly see the differences between values of $\alpha$.  It also shows the difference between CVaR and VaR.  8) $\CVaR_\alpha - \VaR_{0.1}$ when $\alpha$ is radially decaying, with $\alpha=1$ at the center and decaying to $\alpha=0$ at the edges.  Notice that the output risk also decays radially from the center of the map.}
  \label{fig:output_alldata}
\end{figure}

Figure \ref{fig:output_alldata} shows corresponding outputs for the inputs shown in Figure \ref{fig:input_features}.  The leftmost column shows the ground truth traversability cost, while the next 3 columns show CVaR at varying $\alpha$ levels ($\alpha=0.1,0.5,0.9$).  Note that CVaR steadily increases as $\alpha$ increases, and generally is a greater value than the traversability cost.  The next three columns show CVaR at the same varying alpha levels, minus the VaR output of the network when $\alpha=0.1$.  This provides more insight into the changes in VaR and CVaR as $\alpha$ is increased.  The final column shows the same $\CVaR - \VaR_{0.1}$ except with a varying $\alpha$ input, in a radially decaying pattern, with high $\alpha=1$ in the center, and low $\alpha=0$ at the edges.  This kind of pattern could be useful for a risk-aware system which wishes to be more conservative when planning motions in the regions near the robot.  Such a behavior could be useful when the user believes he can rely more on data closer to the robot with greater sensor coverage, as well as wishing to allow for a relaxed commitment to avoiding traversability risks further away.


\subsection{Training}
Next we describe some details of training and hyperparameters.  We used weights for the CVaR loss (\ref{eq:loss}) of $\lambda_V=10.0, ~ \lambda_C=1.0  ~\lambda_m=1.0e-4$.  We used a Huber smoothing coefficient of $h=1.0e-3$.  The network was trained on a $90:5:5$ split of training, validation, and test data.  We used Adam optimization with an initial learning rate of 0.0005, and batch size of 1.  No pre-training of the network was used, weights were initialized with Xavier random weights.  During training, data augmentation was used - including rotation, translation, scaling, shearing, and flipping.  
Training took about 16 hours on a 12GB NVIDIA Tesla K80 GPU.


\section{Evaluation and Results} \label{sec: eval_results}
\begin{figure}
    \centering
    \includegraphics[width=1.0\linewidth]{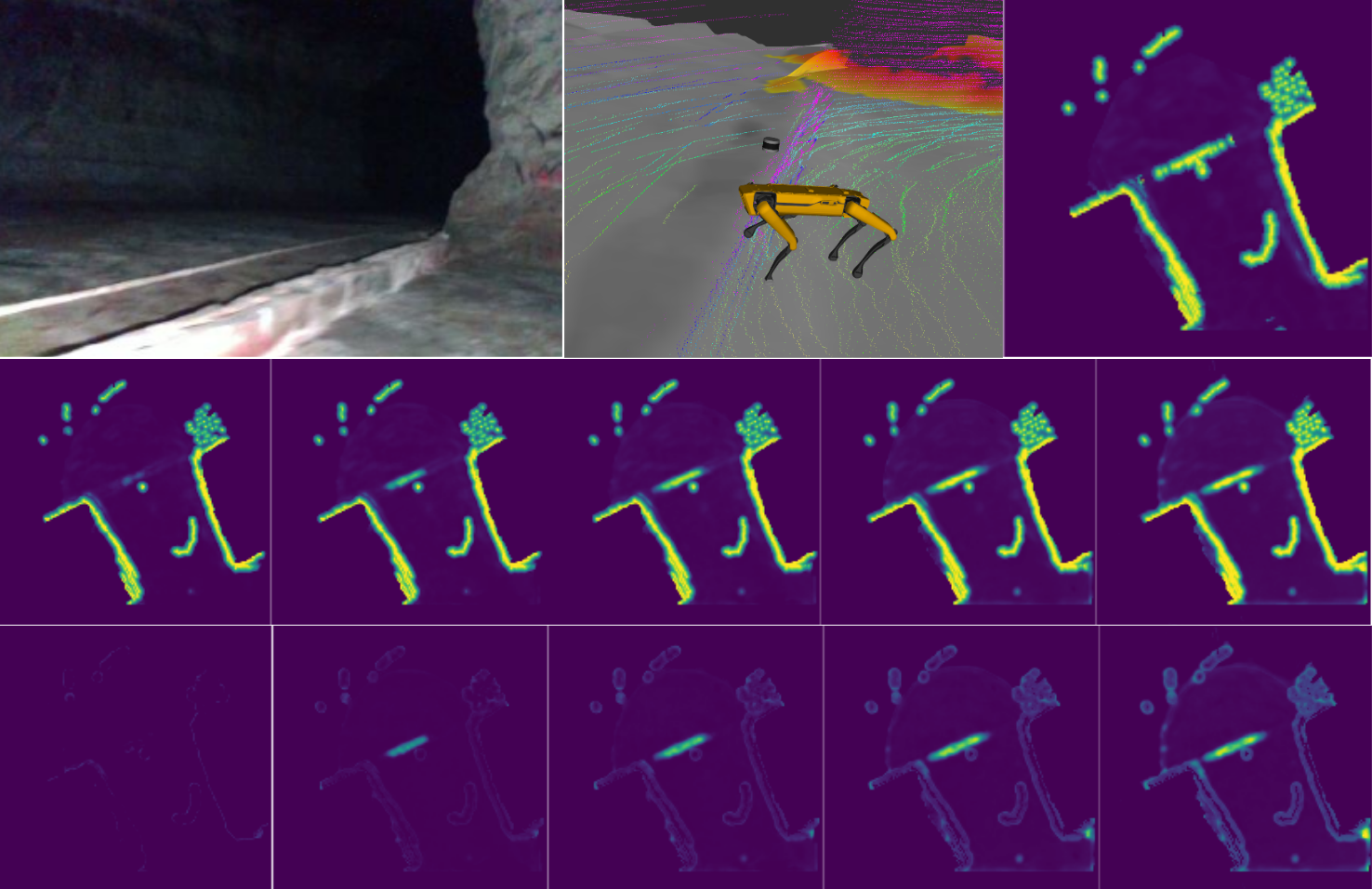}
  \caption{Case study for a low wall obstacle.  Top left:  On-board camera image of the obstacle. Top middle:  pointcloud and elevation map.  Spot has placed one foot on top of the wall.  Top right:  Handcrafted traversability cost.  The wall appears in the center of the map in a line.  Middle row:  CVaR at varying levels of $\alpha=0.1,0.3,0.5,0.7,0.9$ respectively.  Bottom row:  $\CVaR_\alpha - \VaR_{0.1}$ at varying levels of alpha.  The risk of the wall increases greatly as $\alpha$ is increased, while other regions increase in risk more gradually.  Color scale of all maps range from 0 to 1.}
  \label{fig:casestudy_step}
  \vspace{1mm}
\end{figure}

In this section, we evaluate the method on data which is both in-distribution and out-of-distribution, compare against different baselines, and investigate the behavior of the network.  We first present some qualitative observations about the performance of the model.  Figure \ref{fig:casestudy_step} shows an interesting case where the traversability cost of an obstacle is difficult to determine.  In this case (taken from the Limestone Cave dataset), a row of cinderblocks (0.15-0.2m in height) blocks Spot's path.  Spot is usually capable of traversing such obstacles but they do pose some risk.  (In this particular instance, the robot tripped over and fell down.)  We observe the network has learned the concept of the variation of this risk as $\alpha$ is increased from 0.1 to 0.9.  While the handcrafted traversability cost shows a risky region on the wall for this sample, other samples or other similar situations have shown less traversability cost, so the network has learned to smoothly interpolate along this distribution.  Note that this change is more than simple scaling up all costs in the output, as can be seen from the plots of $\CVaR_\alpha - \VaR_{0.1}$.  As $\alpha$ increases, we see a much stronger increase in CVaR at the location of the low wall relative to the rest of the costmap.

We also observe the network learns to remove spurious noise and artifacts from the costmap.  These artifacts occur sporadically, which means they lie far out in the tail of the risk distribution, and are often generated by localization failure or noise.  The network produces a clean risk map which ignores these artifacts.




\subsection{Evaluation Metrics}
We introduce three evaluation metrics for assessing the quality of VaR and CVaR learning for a given value of $\alpha$.  First, we compute the implied-$\alpha$ ($I_\alpha$) value, which is a measure of how closely the predicted VaR value matches the true quantile:
\begin{equation}
    I_\alpha = \frac{\sum_{\mathcal{D}}\mathbb{I}[R <= V_\theta(\alpha)]}{\sum_{\mathcal{D}}\mathbb{I}[mask]}
\end{equation}
An accurate model should have a value of $I_\alpha\approx\alpha$.  Note that we sum over all valid (non-masked) pixels and data samples.  The denominator counts the number of non-masked pixels.

The $R^2$ \textit{coefficient of determination} metric is useful for comparing the explanatory power of a model relative to the distribution of the data.  To quantify the modeling capacity of the network when compared to other baselines, we use a pseudo-$R^2$, which is a function of $\alpha$.  For assessing pseudo-$R^2$ for VaR, we use the following metric \cite{koenker1999goodness}:
\begin{equation}
    R_V^2(\alpha) = 1 - \frac{\sum_{\mathcal{D}} \big[\alpha (R - V_\theta(\alpha))_{+} + (1-\alpha) (R - V_\theta(\alpha))_{-}\big]}
    {\sum_{\mathcal{D}} \big[ \alpha (R - \bar{V}_\alpha)_{+} + (1-\alpha) (R - \bar{V}_\alpha)_{-}\big]}
\end{equation}
where $\bar{V}_\alpha$ is the constant $\alpha$-quantile computed from the training data.  This metric compares the Koenker-Bassett error of a quantile regression model against the total absolute deviation of the data from an independent $\alpha$ quantile.  A value of 1 is the best, reflecting total explanation of the dependent variable, while a value less than 0 is poor, implying worse explanatory power than the simple quantile.

Finally, to assess the CVaR modeling, we construct a similar pseudo-$R^2$ metric for CVaR.  Instead of the Koenker-Bassett error, we use the empirical CVaR error (see \cite{rockafellar2014superquantile} for a rigorous discussion of a similar but computationally less tractable approach).  Our CVaR pseudo-$R^2$ metric compares the total absolute error between our CVaR model and an average empirical CVaR computed from the training data:
\begin{align}
    R_C^2(\alpha) &= 1 - \frac{\sum_{\mathcal{D}} \big|C_\theta(\alpha) - \big(V_\theta(\alpha) + \frac{1}{1-\alpha}(R-V_\theta(\alpha))_{+}\big)\big|}
    {\sum_{\mathcal{D}} \big|\bar{C}_\alpha - \big(\bar{V}_\alpha + \frac{1}{1-\alpha}(R-\bar{V}_\alpha)_{+}\big)\big|}
\end{align}
where $\bar{C}_\alpha$ is the constant $\alpha$-CVaR computed from the training data.

\subsection{In-distribution (ID) Performance}

\begin{figure}
  \centering
  \includegraphics[width=0.49\linewidth, trim=0 0 0 0.75cm, clip]{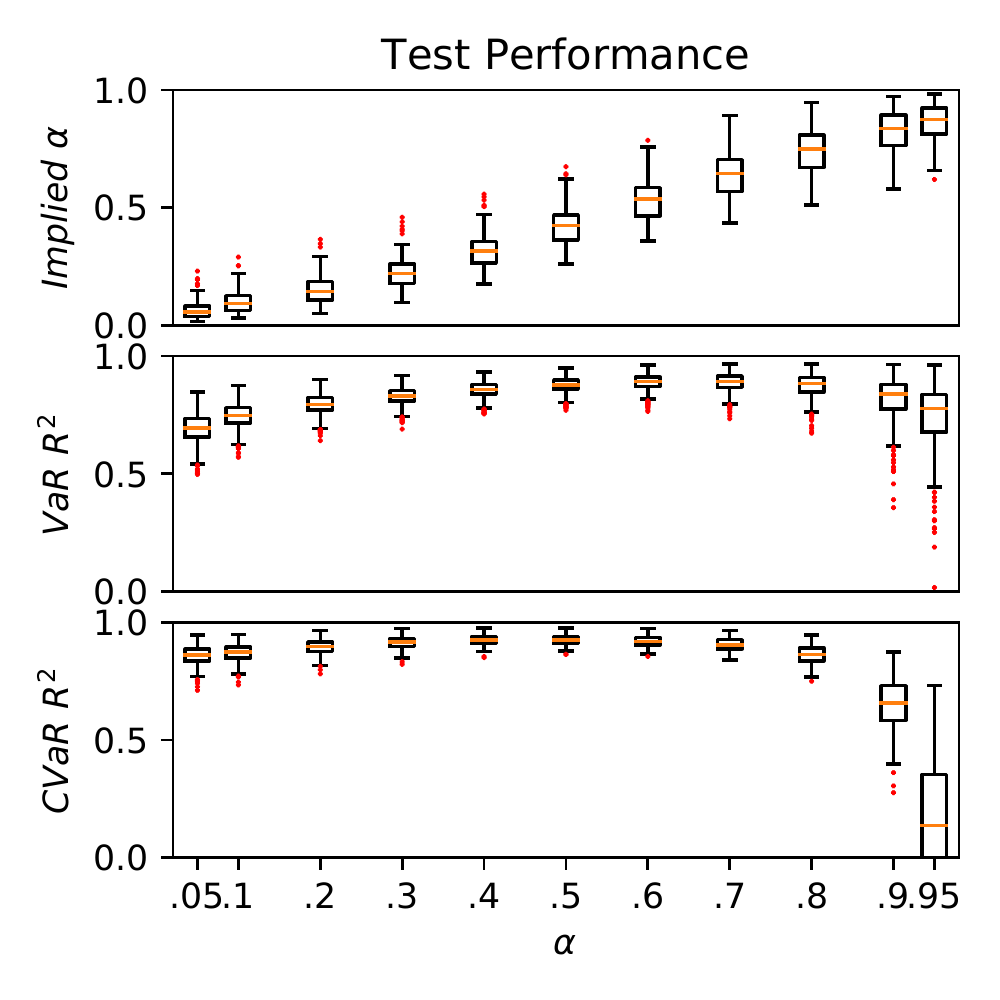}
  \includegraphics[width=0.49\linewidth]{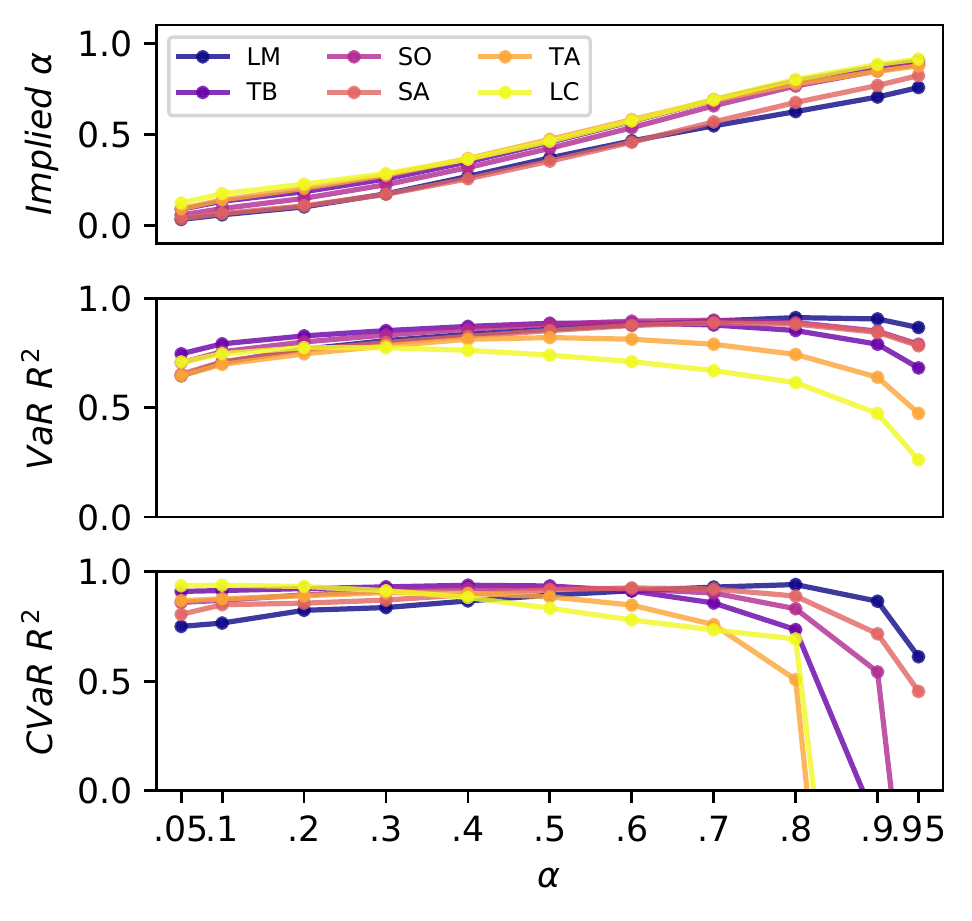}
  \caption{ID performance:  (Left) Boxplot of evaluation metrics for model trained on all 6 datasets, evaluated on held-out test data from all 6 datasets.  The evaluation metrics are computed on each sample individually, and aggregated to make a box plot. (Right) Evaluation metrics for model trained on all 6 datasets, evaluated on held-out test data from 6 datasets individually.}
  \label{fig:indist_alldata}
  \vspace{1mm}
\end{figure}

We begin by evaluating the in-distribution (ID) performance of the network.  Figure \ref{fig:indist_alldata} shows a boxplot of the three evaluation metrics described above, testing on the held-out test data, for a network trained on all 6 datasets.  The box plots show the distribution of metrics computed per-sample.  Implied $\alpha$ holds up well, forming tracking $\alpha$, while VaR $R^2$ and CVaR $R^2$ are close to 1 for most values of $\alpha$.  As $\alpha$ nears 0.9 and 0.95, the CVaR $R^2$ drops closer to 0, implying less of the data is being accurately explained by the model.  This is likely due to the distribution of traversability costs lying between 0 and 1.  When $\alpha$ is high, both VaR and CVaR are more frequently predicted to be 1.  This results in a lower CVaR $R^2$ since the denominator will be small.

We also plot evaluation metrics on each of the 6 dataset categories individually (Figure \ref{fig:indist_alldata}).  This allows us to evaluate the differences between each dataset.  We note that the Limstone Cave dataset has the highest levels of risk, which results in more frequent distortion of the CVaR $R^2$ metric for high $\alpha$ levels.  Nevertheless, all datasets perform well, with pseudo-$R^2$ values close to 1.0.

\subsection{Out-of-Distribution (OOD) Performance}
\begin{figure}
  \centering
  \includegraphics[height=0.5\linewidth, trim=0 0 0 0.75cm, clip]{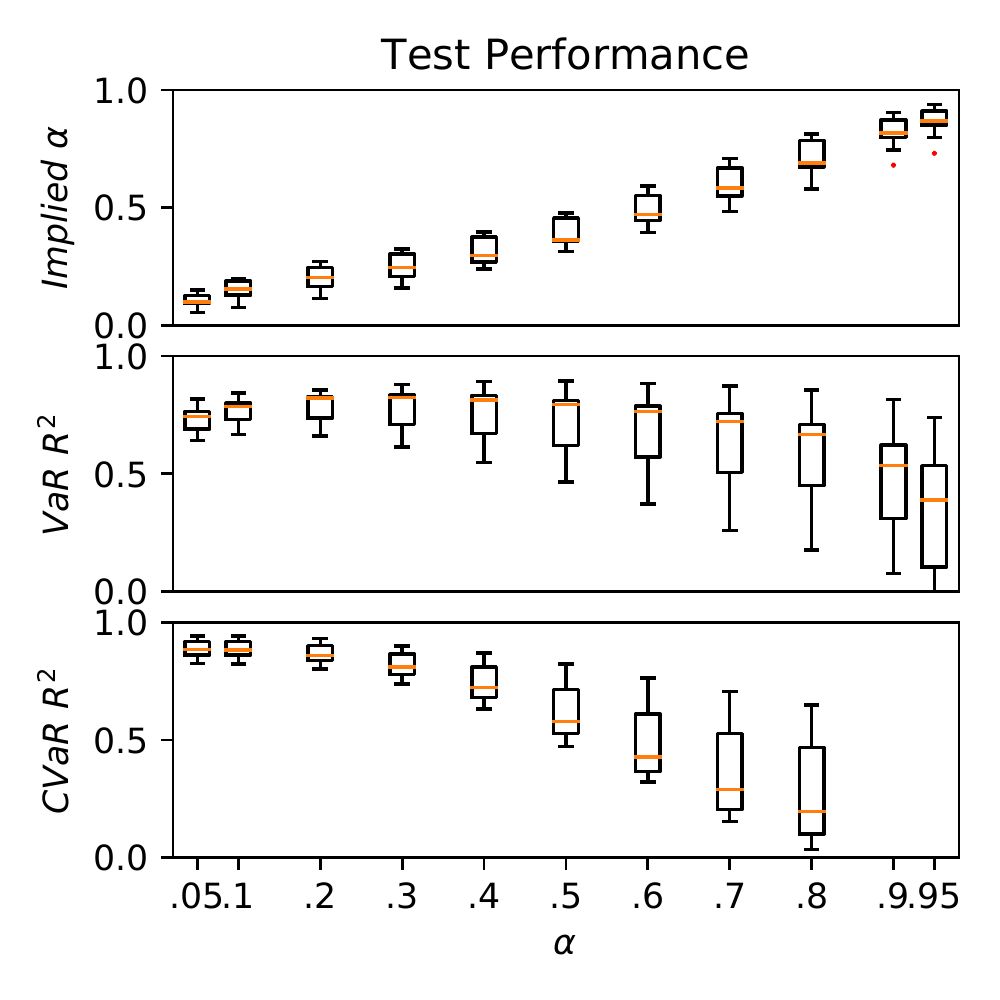}
  \includegraphics[height=0.5\linewidth, trim=1.75cm 0 0 0.75cm, clip]{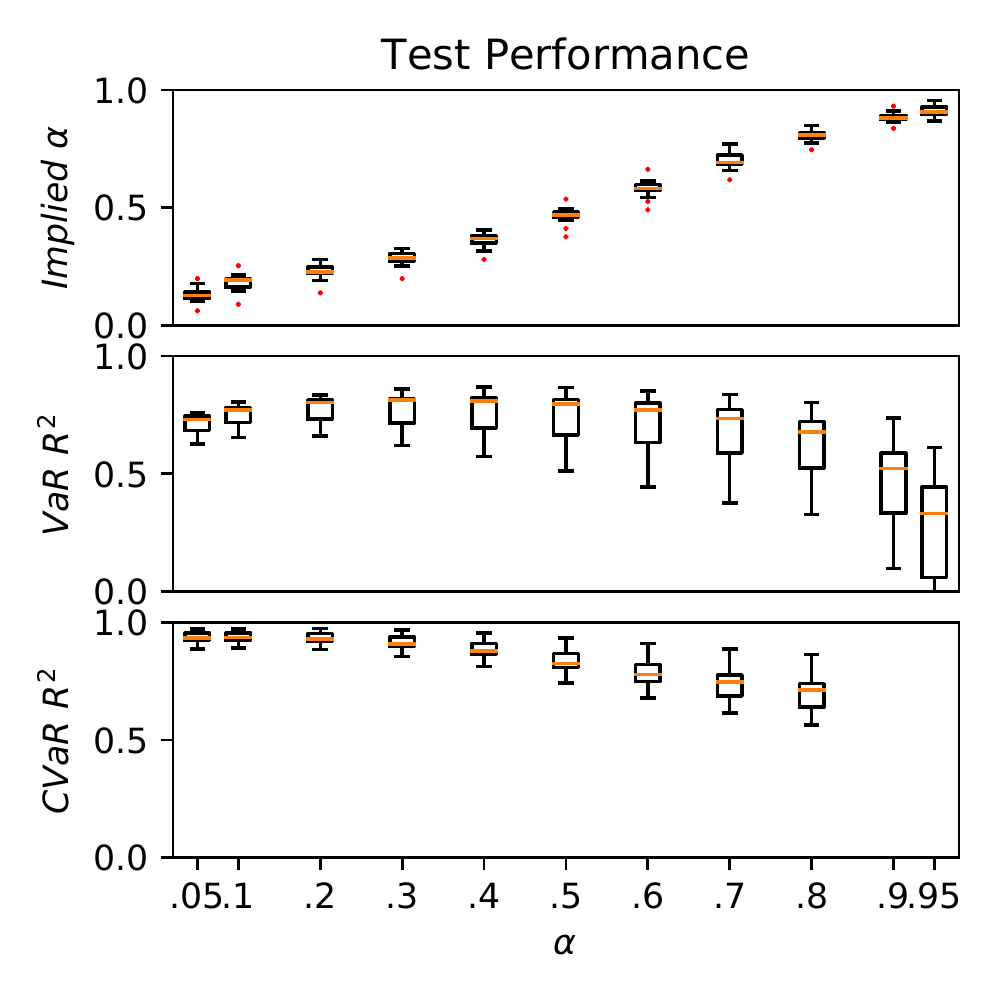}
  \caption{OOD performance:  (Left) Boxplot of evaluation metrics for model trained on all datasets except the Limestone Cave, evaluated on held-out data from the Limestone Cave.  Performance is worse than the ID performance, i.e. when the model is trained on all data, and evaluated on the Limestone Cave data (Right).}
  \label{fig:outdist_heldout_cave}
\end{figure}

To evaluate out-of-distribution (OOD) performance, we train the network on all data except the one dataset, and then evaluate the network on the OOD dataset (which the network has not seen during training).  We do this by holding out the Limestone Cave dataset (Figure \ref{fig:outdist_heldout_cave}).  The Limestone Cave dataset has the most different and challenging terrain with the highest average risks, and with many features not present in the other datasets, such as sharp jagged rocks, narrow passages, channels of water, and rough floors.  We see that OOD performance on this dissimilar dataset desires improvement, in contrast to OOD performance on other more similar datasets (not shown).  However, the Limestone Cave OOD model is still able to maintain healthy pseudo-$R^2$ statistics ($>0$) for a large range of $\alpha$.


\subsection{Comparisons against baselines}

\begin{figure}
  \begin{minipage}[c]{0.5\linewidth}
    \includegraphics[width=\linewidth]{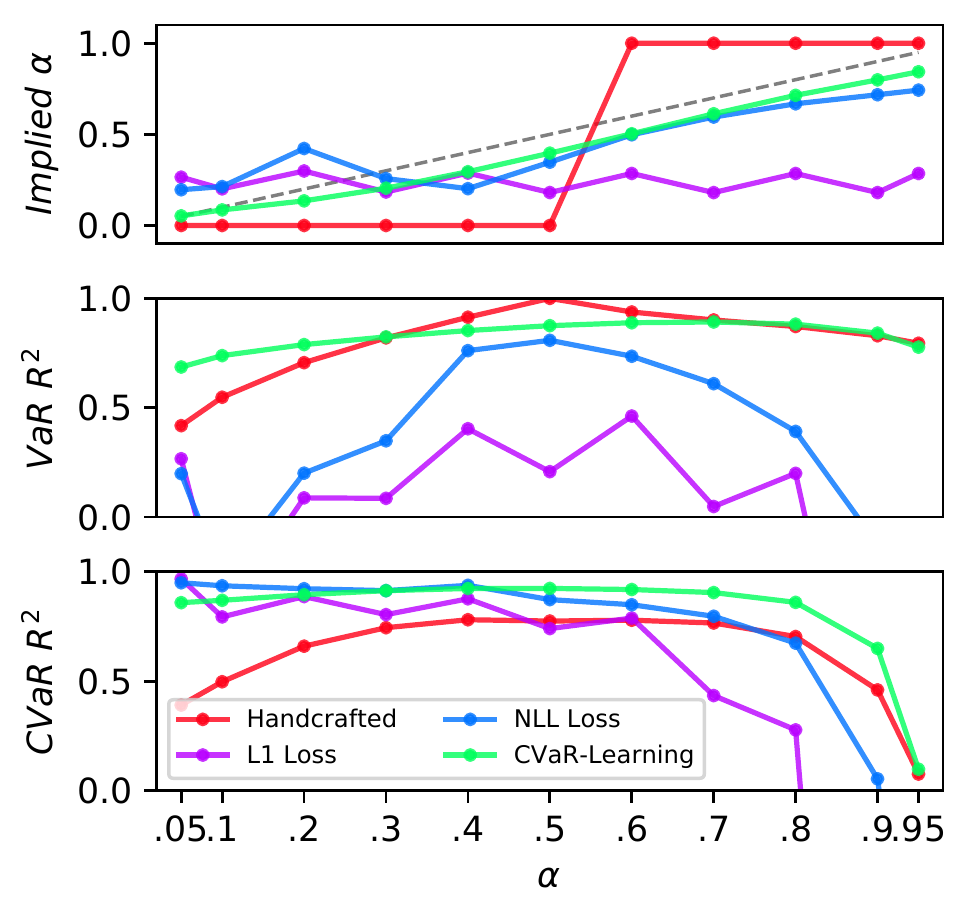}
  \end{minipage}%
  \begin{minipage}[c]{0.48\linewidth}
    \caption{Comparing CVaR-learning model against handcrafted CVaR cost model, L1-loss model, and NLL-loss model.  The learned model was trained on all 6 datasets.  Shown here are evaluation metrics computed over the held-out test data for all 6 datasets.  The CVaR-learning method outperforms by a clear margin.  Dotted line in first plot indicates ideal implied $\alpha$ values.
    } \label{fig:baseline_compare}
  \end{minipage}
\end{figure}

Finally, we compare our approach against three baselines:  1) the handcrafted traversability risk model which relies on assumptions of Gaussianity and accurate mean and variance, described in Section \ref{sec:handcraft}, 2) a model trained to predict the traversability cost using an L1 loss function only, and 3) a model trained to predict both traversability mean and variance using a Gaussian negative log likelihood (NLL) loss \cite{Kendall2017}, which is then used to compute VaR and CVaR assuming a Gaussian distribution.  We computed metrics by taking the average over all test datasets, shown in Figure \ref{fig:baseline_compare}.  The L1 Loss model does not capture any quantile information and does not depend on $\alpha$.  Therefore, it does not capture the VaR or CVaR as well as the trained model.  The hand-crafted traversability cost model also does not perform well - it fails to accurately capture the quantile (VaR) for most values of $\alpha$.  It also does not provide as accurate CVaR values as the learned model.  The NLL loss model fairs relatively well predicting CVaR, but less-so for VaR.  However, the learned CVaR model still outperforms it in the CVaR statistic.

We also consider the computation time of using the handcrafted traversability cost method vs. our CVaR learning method (Table \ref{table:computation_time}).  The CVaR learning approach tends to be almost 5x more efficient then the handcrafted approach.  (Both methods currently rely on ground segmentation and elevation mapping steps.)  We see that replacing the handcrafted costmap with the learned one removes the largest bottleneck in the traversability assessment pipeline.

\begin{table}[h!]
\vspace{1mm}
\centering
\small
\begin{tabular}{c|c|c} 
  Computation Time & $\mu$ (ms) & $\sigma$ (ms)\\
  \hline
 Ground Segmentation & 13.97 & 3.88\\
 Elevation Mapping & 140.19 & 65.48\\
 \hline
 Handcrafted Costmap & 241.92 & 30.11\\
 CVaR-Learning Costmap & \textbf{48.94} & \textbf{15.36}
\end{tabular}
\vspace*{0.1cm}
\caption{Computation time for handcrafted costmap vs. CVaR-learning costmap, computed from N=250 samples.  Also shown are times for preprocessing steps, i.e. ground segmentation and creating the 2.5-D elevation map.}
\label{table:computation_time}
\end{table}


\section{Conclusion}
In this work, we introduced a novel neural network architecture for transforming pointcloud data into risk-aware costmaps.  We also introduced a novel formulation of CVaR and VaR loss functions which train the network to accurately capture tail risk, without relying on cumbersome modeling or distribution assumptions.  We demonstrated this approach on a robotic navigation task in unstructured and challenging terrain.  We evaluated the method using metrics which show that this approach reliably learns the expected tail risk given a desired probability risk threshold between 0 and 1 and produces a traversability costmap which is robust to outliers, accurately captures tail risks, generalizes well to out-of-distribution data, and is more computationally efficient. Future work includes investigation of architectures which directly map LiDAR points to costs, investigating the use of importance sampling to improve risk estimates for high values of $\alpha$ \cite{deo2020optimizing}, and investigation of learning dynamic risk metrics \cite{majumdar2020should}.

\section{Acknowledgement}
We thank Keuntaek Lee, Kyohei Otsu, Anushri Dixit, and Nicholas Palermo for useful discussions and contributions to the code base.  This research was partially carried out at the Jet Propulsion Laboratory (JPL), California Institute of Technology, and was sponsored by the JPL Year Round Internship Program and the National Aeronautics and Space Administration (NASA).  
\bibliographystyle{abbrvnat}
\bibliography{main}

\end{document}